\documentclass[10pt]{article}

\usepackage[margin=1in]{geometry}
\usepackage[utf8]{inputenc}
\usepackage[T1]{fontenc}
\usepackage{lmodern}
\usepackage{microtype}
\usepackage{booktabs,tabularx}
\usepackage{graphicx,xcolor}
\usepackage{amsmath,amsfonts,amssymb}
\usepackage[numbers,sort&compress]{natbib}
\usepackage[colorlinks=true,allcolors=blue]{hyperref}
\usepackage[font=small,labelfont=bf]{caption}
\newcolumntype{L}{>{\raggedright\arraybackslash}X}
\title{Paris as a 15-Minute City: An Explainable AI Perspective\thanks{This manuscript is the detailed report of a poster presented at NetMob 2025, 8 October 2025.}}

\author{Andr\'as J. Moln\'ar \and Csaba I. Sidl\'o\thanks{Corresponding author: \texttt{sidlo@sztaki.hu}} \and Rita R\'onai \and Domonkos R\'ozsay\\[0.5em]
\small Institute for Computer Science and Control, Hungarian Research Network, Hungary}
\date{}

\begin{document}

\maketitle

\begin{abstract}
The 15-minute city promotes access to everyday services within a short walk or bicycle ride, but its relationship with observed mobility remains difficult to quantify.
We investigate this relationship in the Paris metropolitan area using mobility trajectories from the NetMob 2025 Data Challenge, enriched with INSEE sociodemographic data and OpenStreetMap points of interest (POIs), yielding approximately 70,000 trip segments after stop-based segmentation and data cleaning.

We construct walking- and cycling-based indicators of local service availability and examine their associations with trip duration, transport mode, and short-trip car use. 
Higher POI availability is associated with less private motorized travel and more active mobility, although this relationship is substantially weaker in the outer agglomeration. 
Gradient-boosted tree models interpreted with explainable machine-learning methods consistently identify trip purpose, home--work distance, local service availability, vehicle ownership, public-transport subscription, and sociodemographic context as important predictors.
For short trips, high POI density is associated with lower car use, while car ownership and driving-licence availability are associated with higher predicted car use; where services are sparse, public-transport subscription is associated with lower predicted car dependence. 
Finally, explainable AI (XAI) methods are used to examine how feature attributions change under alternative assumed variable orderings.

The results are consistent with central assumptions of the 15-minute city while revealing substantial spatial and demographic heterogeneity. 
They also demonstrate how explainable machine-learning methods can complement accessibility indicators and identify locally relevant hypotheses for urban-mobility policy.
\end{abstract}

\bibliographystyle{plainnat}

\section{Introduction and Related Work}

The need for climate action, better urban livability, and sustainable mobility has revived interest in concepts reimagining modern cities.
The 15-minute city, a concept emphasizing human-scale urbanism where essential services are within a 15-minute walk or bicycle ride, promotes decentralization, strong communities, and reduced reliance on private motorized transport \cite{moreno15minutecity, weng201915, pozoukidou15minutes, abdelfattah2022} while advancing equity, sustainability, and well-being \cite{dutPosition2024, dutMapping2024}. 

Diverse approaches operationalize the 15-minute city:  
Knap et al.\ introduce a composite cycling accessibility index to identify spatial and socioeconomic inequalities in Utrecht \cite{KNAP2023100043}, while Abbiasov et al.\  use mobility datasets from global cities to measure disparities in access to essential services \cite{abbiasov2022mobility15minutes}.
Paris stands out as a flagship case, with Thaury et al.\ analysing its accessibility patterns to assess alignment with the 15-minute framework \cite{thaury2024city}.

The 15-minute city faces criticism and implementation challenges: concerns include social surveillance, restrictions on individual mobility, and doubts about feasibility in car-oriented or socioeconomically segregated areas \cite{caprotti2024, mouratidis2024, ube2023}. 
The adaptability to diverse urban contexts raises equity issues, as critics warn that without inclusive planning and targeted investment, the model could reinforce existing inequalities \cite{szymanska2024, smartcities2023, mouratidis2024}.

This paper evaluates the 15-minute city concept through correlation analysis and explainable artificial intelligence methods. It further uses asymmetric SHAP to examine how model attributions change under alternative assumed variable orderings. We assess whether empirical evidence supports the principles of the 15-minute city by integrating the NetMob 2025 dataset \cite{netmob25} with geographic and sociodemographic data \cite{insee}. Our aim is to examine the extent to which observed mobility patterns align with central assumptions of the 15-minute city concept. We analyse how POI availability is associated with travel-mode choices and sociodemographic characteristics, highlighting the interplay among urban mobility, population characteristics, and metropolitan spatial structure. We also illustrate how explainable machine-learning methods can be used to extract interpretable patterns from the combined data.




\section{Data Preparation and Analytical Methodology}

\subsection{External Data}
NetMob 2025 trajectories and demographic data were enriched with the latest available 2020 INSEE demographics \cite{insee} 
and October 2022 OpenStreetMap (OSM) data provided by the Ohsome API \cite{OpenStreetMap, ohsomedb}. 

Approximate home and workplace locations were determined for the surveyed persons by the starting and/or ending points of their respective trips. OpenStreetMap data of various POI categories was linked to these locations, and a proximity-based POI availability indicator was calculated using circular buffers corresponding approximately to 15 minutes of walking and cycling\footnote{We used circular buffers rather than street-network routes for this initial analysis. The resulting indicators should therefore be interpreted as proximity-based accessibility proxies rather than exact travel-time measures.}. See Table \ref{tab:poi-tags} for the POI types we formed and the actual tags representing particular POI categories we queried and grouped into these types.

\begin{table*}[t]
\centering
\caption{Points of interest (POI) types and their OSM tags selected for quantifying 15-minute POI availability at queried locations}
\label{tab:poi-tags}
\begin{tabularx}{\linewidth}{l L}
\toprule
\textbf{POI type} & \textbf{POI tags} \\
\midrule
education &
\verb|amenity=kindergarten|, \verb|amenity=school|, \verb|amenity=training|, 
\verb|amenity=driving_school|, \verb|amenity=language_school|, \verb|amenity=music_school|, \verb|amenity=library|
\\[0.35em]

healthcare &
\verb|amenity=pharmacy|, \verb|amenity=hospital|, \verb|amenity=dentist|, \verb|amenity=veterinary|, 
\verb|emergency=defibrillator|, \verb|shop=optician|, \verb|amenity=social_facility|, \verb|social_facility=nursing_home|
\\[0.35em]

municipal &
\verb|leisure=playground|, \verb|leisure=park|, \verb|leisure=garden|, 
\verb|amenity=arts_centre|, \verb|tourism=museum|, \verb|tourism=gallery|, \verb|amenity=marketplace|, 
\verb|amenity=community_centre|, \verb|amenity=fire_station|, \verb|amenity=police|
\\[0.35em]

finance\_and\_telecomm &
\verb|amenity=bank|, \verb|amenity=post_office|, \verb|amenity=atm|, \verb|amenity=bureau_de_change|
\\[0.35em]

commercial &
\verb|amenity=restaurant|, \verb|amenity=bar|, \verb|amenity=cafe|, \verb|amenity=fast_food|, 
\verb|amenity=bakery|, \verb|amenity=butcher|, \verb|shop=convenience|, 
\verb|shop=supermarket|, \verb|shop=department_store|, \verb|shop=clothes|, \verb|shop=shoes|, 
\verb|shop=beauty|, \verb|shop=hairdresser|, \verb|shop=furniture|, \verb|shop=electronics|, 
\verb|shop=stationery|, \verb|shop=tobacco|, \verb|shop=books|, 
\verb|amenity=cinema|, \verb|amenity=theatre|, \verb|amenity=nightclub|, 
\verb|leisure=fitness_centre|, \verb|sport=multi|, \verb|tourism=hotel|
\\[0.35em]

accessibility &
\verb|amenity=taxi|, \verb|amenity=bicycle_parking|, \verb|amenity=bicycle_repair_station|, 
\verb|amenity=fuel|, \verb|amenity=parking|, \verb|public_transport=stop_position|, \verb|public_transport=station|
\\
\bottomrule
\end{tabularx}
\end{table*}

\subsection{Trips and Trip Segments}
We experimented with three trip-classification schemes when analysing whole trips: 1) duration-based, whether the trip is actually under 15 minutes or more; 2) a three-way classification, separating walking and cycling trips from public transport and private motorized trips; 3) a two-way classification based on trip modes and duration, where the first class consists of all walking and cycling trips as well as those public transport trips which remain under 15 minutes, and the second class contains all private motorized trips and longer public transport trips.

To better capture the structure of individual movements, we refined trip trajectories by applying stop detection, a common pre-processing operation. 
Many recorded trips contained intermediate pauses that effectively split them into meaningful trip segments. 
Using MovingPandas \cite{graser2019movingpandas}, we defined a stop as remaining within a 100~m radius for at least 15 minutes. We also experimented with a 50~m radius. 
The trip segments between stops could then be used as the basis for subsequent analysis. 
This segmentation provides a more detailed representation of the recorded movement structure and supports the calculation of segment-level characteristics such as average speed and travel distance.


\subsection{POI-based Features}
Proximity-based ``15-minuteness'' indicators were constructed by counting six POI categories (education, healthcare, municipal services, finance/telecom, commercial, transport) within a 1.2 km walking radius of home/work locations, and 4 km cycling radius, respectively. 
These indicators may be refined in future work by adjusting POI types, radius, or routing (see e.g. \cite{mouratidis2024}).   

OSM tagging practices can cause a single facility to be represented by multiple geometrical elements. Consequently, raw OSM feature counts do not always correspond directly to the number of distinct real-world facilities. We therefore applied a logarithmic transformation to each POI category to reduce the influence of extreme counts while preserving differences among low-count areas. This transformation mitigates skew but does not explicitly deduplicate facilities; the resulting indicators should therefore be interpreted as approximate measures of local POI availability.  



\subsection{Statistics and Unsupervised Learning}

For the statistical analysis, categorical demographic features were binarized and all numerical features were min--max normalized.

We first calculated pairwise Pearson correlations to identify associations among variables and groups of related features.

The correlation analysis was carried out in three steps: 1) correlation among different POI types, 2) correlation of certain trip characteristics with 15-minute POI availability, 3) a user-level correlation matrix, with variables of their demographics, and aggregate values of their trips. For 3) we built a correlation matrix and removed redundant or weak variables. 
For example, we kept total trip time instead of mode-specific counts, while retaining leisure trips as a quality-of-life indicator. The final feature set reflected the most meaningful correlations.

\subsection{Explainable AI Models}
Explainable AI (XAI) techniques offer insights into model behaviour \cite{longo2024explainable}, enabling interpretation of 15-minute city--related target variables. We employed gradient-boosted trees for their strong predictive performance and compatibility with interpretability frameworks such as SHAP (SHapley Additive Explanations) \cite{lundberg2017shap}.

Gradient-boosted tree models such as XGBoost capture nonlinear relationships and feature interactions. Their predictions can be interpreted using SHAP, while individual trees can illustrate local feature combinations represented within the ensemble. Such explanations may help formulate policy-relevant hypotheses, although they do not by themselves establish the effects of possible interventions. 

For example, if a model associates limited bicycle infrastructure with lower bicycle use in areas containing many short trips, the result may motivate further local investigation of whether additional infrastructure would be beneficial. Any intervention would require independent validation and consideration of the broader urban context.

\subsection{Asymmetric SHAP Model Explanations} 
Beyond conventional SHAP analysis, we use asymmetric Shapley values to examine how feature attributions change when selected groups of variables are constrained by an assumed partial ordering. This analysis does not identify causal relationships from the data. Rather, it provides a sensitivity analysis of model explanations under alternative domain-informed causal hypotheses, following the asymmetric attribution framework of Kelen et al.~\cite{kelen2022causal}.


\section{Results and Discussion}
Statistical analysis, along with preliminary unsupervised and supervised machine-learning models, revealed several patterns relevant to the 15-minute city concept in the Paris dataset. The analysis focused on aspects closely related to the 15-minute city concept, aiming to assess whether observed mobility patterns are consistent with mobility and accessibility outcomes commonly associated with the 15-minute city concept. The available data do not provide a direct measure of quality of life or livability. A central objective of this investigation was to understand how residents use the city and whether their mobility patterns align with the core hypotheses of the 15-minute city model.

Identifying variables associated with mobility outcomes can support the formulation of hypotheses for urban-planning research. By examining the interplay between city structure, sociodemographics, and urban mobility, this research aims to provide insights into how urban-planning questions could be further investigated. In the following sections, we highlight some of these patterns. Nevertheless, many other relevant questions and patterns remain to be explored. The NetMob dataset is particularly useful in this regard, as it offers a rich foundation for further research.

\subsection{Basic Statistics}

We identified approximate home locations for 3,295 and work locations for 2,089 of the 3,337 participants. A total of 2,408 stops were identified in the GPS trajectories, resulting in 1.22 segments per trip on average. After data cleaning and outlier removal, 70,359 valid trip segments remained. 

Although the complete dataset contains a substantial number of segments, some task-specific subsets, such as bicycle trips and short car trips, are considerably smaller. Caution is therefore needed when interpreting model results for specific subsets like 15-minute trips or bicycle commutes. These limitations mean that while the observed patterns are informative, a larger, more comprehensive dataset would be necessary to draw more definitive and broadly applicable conclusions.



\subsection{Correlation Analysis}

\begin{figure}[ht]
\centering
\fbox{\includegraphics[width=150pt]{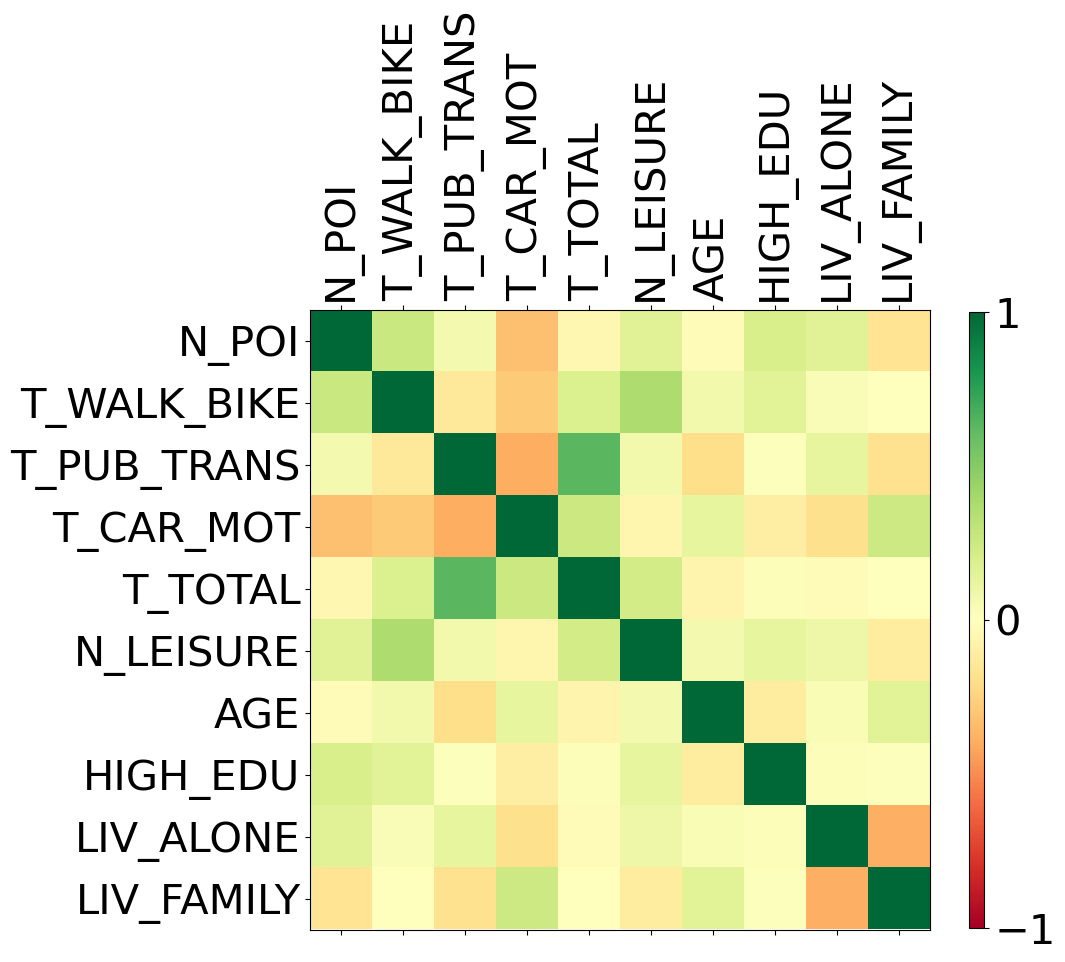}}
\caption{Pearson correlations of important features: 15-minute POIs, total and trip-grouped travel times, number of leisure-aimed trips, age, higher education (4-5 years or more) and living arrangement (alone or with family)}
\label{fig:correlation-matrix}
\end{figure}

Correlations (Figure \ref{fig:correlation-matrix}) among POI categories -- as a first step -- within a 15-minute radius were strong, with minimum pairwise scores of 0.79 (excluding municipal services) and 0.67 (including them). Therefore, a unified POI count was adopted as the core '15-minuteness' indicator for many of our analyses. Expanding the radius to a 15-minute cycling distance did not substantially change the results.

\begin{figure*}[ht]
\centering
\fbox{\includegraphics[width=0.98\linewidth]{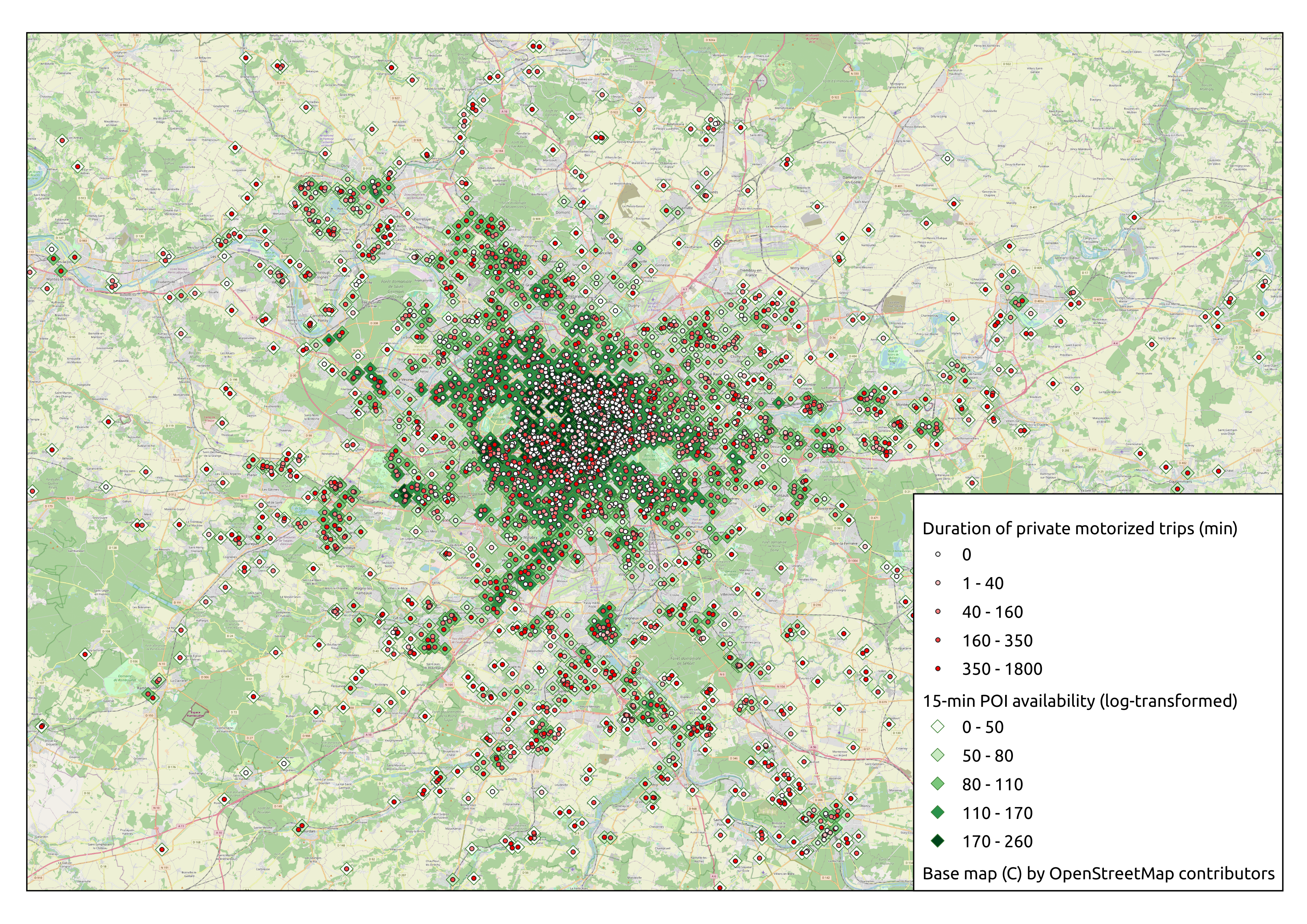}}
\caption{Correspondence map between the duration of all private motorized trips per person and the 15-minute POI availability}
\label{fig:paris-15min-map}
\end{figure*}

\begin{figure*}[ht]
\centering
\fbox{\includegraphics[width=0.98\linewidth]{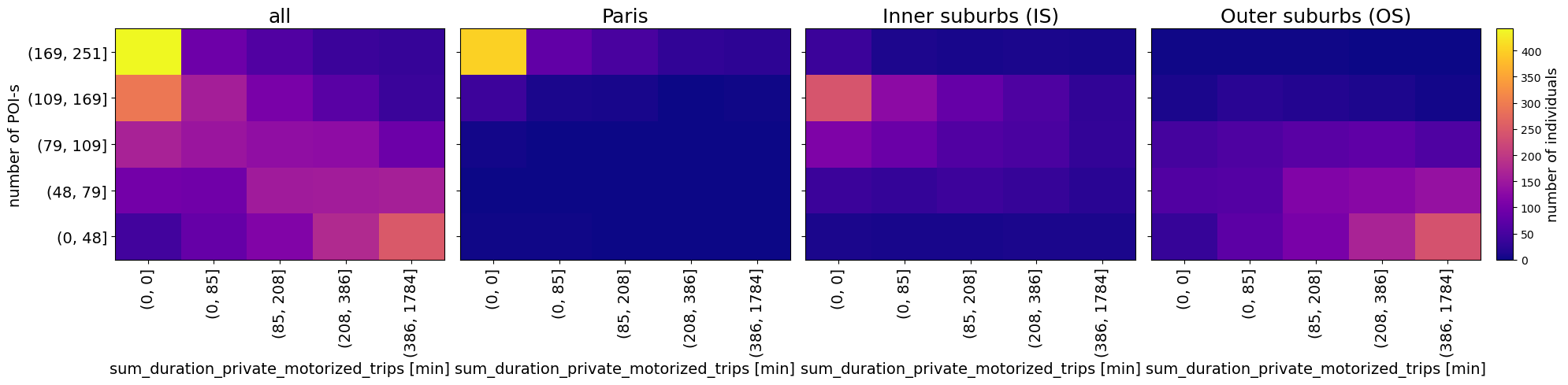}}
\caption{Relationship between 15-minute POI availability and private motorized travel, shown by metropolitan zone.}
\label{fig:poi-sum-dur-priv}
\end{figure*}

\begin{figure*}[ht]
\fbox{\includegraphics[width=0.98\linewidth]{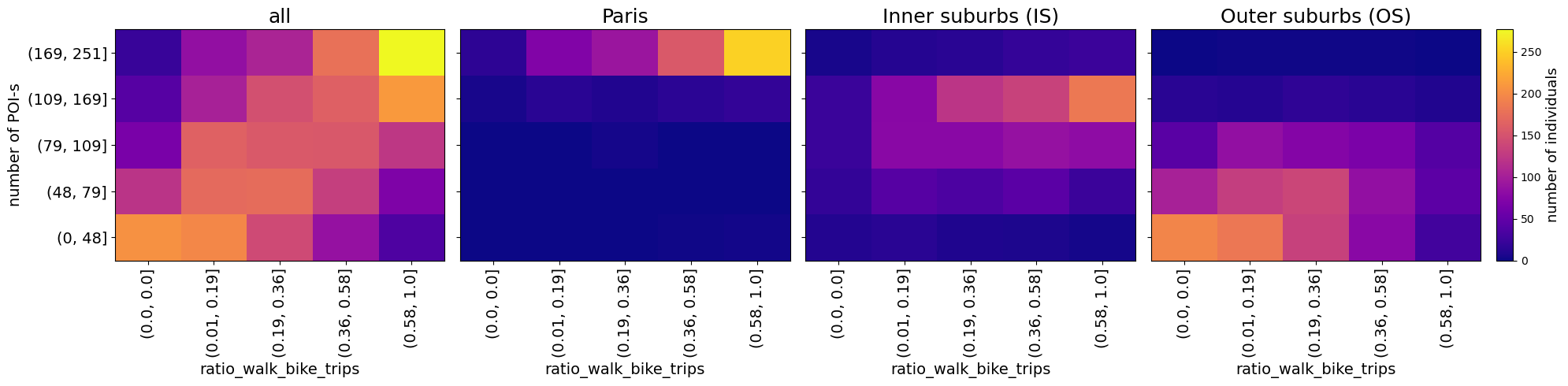}}
\caption{Relationship between 15-minute POI availability and the proportion of trips made by walking or cycling.}
\label{fig:poi-ratio-walk-bike}
\end{figure*}

In the second step, we examined associations between the POI-based ``15-minuteness'' indicator and aggregated user-level trip characteristics. First, a map was constructed (see Figure \ref{fig:paris-15min-map}) for a general overview, using the total time spent for private motorized trips per user. Based on the observations of the map and general intuition we decided to perform this analysis for each of the three location categories Paris, IS (inner agglomeration), OS (outer agglomeration) (see Figure \ref{fig:poi-sum-dur-priv}). Besides car use we also considered the ratio of walking or cycling trips to the total number of trips per user (Figure \ref{fig:poi-ratio-walk-bike}). 

Paris shows a pattern broadly consistent with the 15-minute city concept: POI availability is negatively associated with car use and positively associated with walking and cycling. However, the result may be biased by the area having generally more POI availability than other areas, as well as by restrictive policies on car use. POI availability is substantially lower in the agglomeration, but especially in the inner agglomeration belt it is negatively associated with car use and positively associated with walking and cycling. It is worth noting that the majority here does not use cars at all. One possible explanation is the relative convenience of public transport for journeys into Paris, although this mechanism was not analysed separately. Car use (private motorized trips) is prevalent, however, in the outer agglomeration and even in the presence of POIs a significant number (and duration) of trips are taken by private cars, so the 15-minuteness association is much weaker in these neighbourhoods than in more central locations.

In the third step of our correlation analysis, multiple variables were taken into account. 
Here, 
the three-way trip categorization offered clearer insights: public transport time negatively correlated with private motorized trips (suggesting a 'car user'/'public transport user' divide), and walking and cycling showed a weaker negative correlation with motorized trips.
  

Private motorized trips showed a negative correlation with POI proximity, whereas walking or cycling demonstrated the strongest positive correlation, which is consistent with central assumptions of the 15-minute city concept. Public transport was predominantly used by younger individuals, suggesting positive trends across age groups, and strongly correlated with total travel time.

Leisure trip frequency positively correlated with POI proximity, and highly mobile individuals who walk or cycle took more leisure trips. This result does not provide evidence that POI proximity directly increases free time and indicates that the relationship between accessibility, leisure travel, and livability requires further study.


Household type showed the strongest associations with mobility and POI proximity, followed by age and education. While living alone or holding a degree correlated mildly positively with POI proximity (consistent with inequity critiques), families showed slight negative correlations with POI proximity, public transport, and leisure travel, plus higher car use, highlighting design challenges for their needs.


%


\subsection{Understanding Trip Duration Factors}

Trip duration is an important characteristic of urban mobility and is often considered in discussions of accessibility and livability. 
Since direct metrics for quality of life, well-being or livability were not available for this study (or were not granular enough), we used trip duration and related mobility characteristics as indirect outcomes relevant to the 15-minute city concept. 

Several XGBoost models were trained and evaluated to predict the duration of trip segments, using descriptive features of the trips themselves, along with sociodemographic and 15-minute POI accessibility features of participant's homes and the segment start points. 
We deliberately excluded features that were either closely related to or irrelevant for segment length, such as the names of origin and destination areas, as they do not help in building generic patterns. We also omitted most statistics from a user's other trips, as while they could help create user profiles, they were not considered interesting factors in this case.

\begin{figure}[ht]
\fbox{\includegraphics[width=0.98\linewidth]{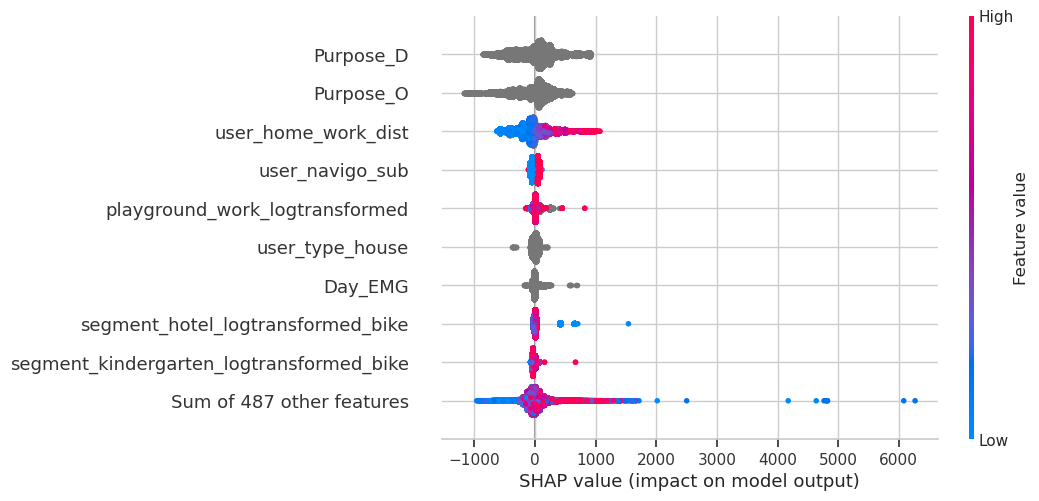}}
\caption{SHAP beeswarm plot with the most important input features and their impact on an XGBoost model trained to predict segment length}
\label{fig:shap-segment-length}
\end{figure}

Depending on model complexity, the models have limited predictive accuracy: test-set $R^2$ ranged from 0.21 to 0.35, with mean absolute errors of approximately 400--700 seconds. 
Figure \ref{fig:shap-segment-length} illustrates the top features of a segment duration XGBoost model, one with 30 trees, maximum depth 3, visualizing the impact of each feature on a model's output. 
Each dot represents a single prediction, with its horizontal position indicating the feature's effect on the output (segment length) and its color showing the feature's value (high or low). Grey rows are categorical features. 

Features derived from all three data sources---trip characteristics, INSEE sociodemographics, and OSM POIs---appear among the higher-ranked model inputs. 

Purpose of the trip is the most important feature according to the models, along with home--work distance: the larger the distance, the higher the predicted trip duration. Similarly, the models predict longer trips for users with public-transport subscriptions. 

Several POI-based features also appear among the ten highest-ranked inputs. For example, hotels, kindergartens and playgrounds (reachable by foot or by bicycle within 15 minutes) near segment starts or workplaces are important factors. 

$Day\_EMG$ is the day of week; weekends and the first day of the week behave differently regarding trip length.

Several sociodemographic features also appear among the higher-ranked inputs, such as $user\_type\_house$, which is the household type of the user: having children in the household is associated with lower predicted segment durations (however, combining with the purposes may change this; going to school, for example, differs from going to work). 

The most important INSEE feature that did not rank among the ten most important features is $DISP\_S80S20S20$, which is the "S80/S20" ratio of disposable income. In this model, higher values of the local disposable-income S80/S20 ratio are associated with lower predicted trip durations, conditional on the other included features. This result should not be interpreted as a causal relationship.

\subsection{Boundaries of X-Minute Mobility}
Within the concept of X-minute mobility, the definition of a "15-minute city" can vary, including different time frames like 15 or 20 minutes and different modes of transport like walking or cycling. 
By analysing trip segments, it is possible to examine how observed travel behaviour changes with the speed and length of trips. The choice of transport mode is a crucial aspect of this behaviour. 

\begin{figure}[ht]
\centering
\fbox{\includegraphics[width=300pt]{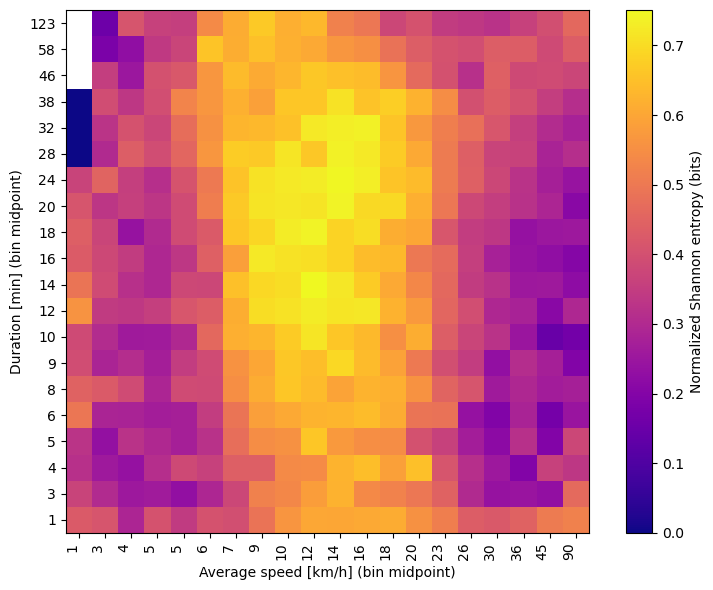}}
\caption{Diversity of transport-mode choices, measured by normalised Shannon entropy, by trip-segment duration and average speed. Higher values indicate more diverse mode choices.}
\label{fig:modes-entropy}
\end{figure}

Figure \ref{fig:modes-entropy} visualizes the diversity of transport-mode choices based on trip-segment duration and average speed. The speed range between 5 and 20 km/h shows a high degree of diversity in transport choices. The observed changes around 5~km/h and 15~km/h are consistent with commonly used walking- and cycling-speed assumptions. While there are no clear boundaries for duration, the common 15-minute choice is situated in the middle of the most variable region, providing empirical support for its use as a practical convention in this dataset.

\begin{figure}[ht]
\centering
\fbox{\includegraphics[width=300pt]{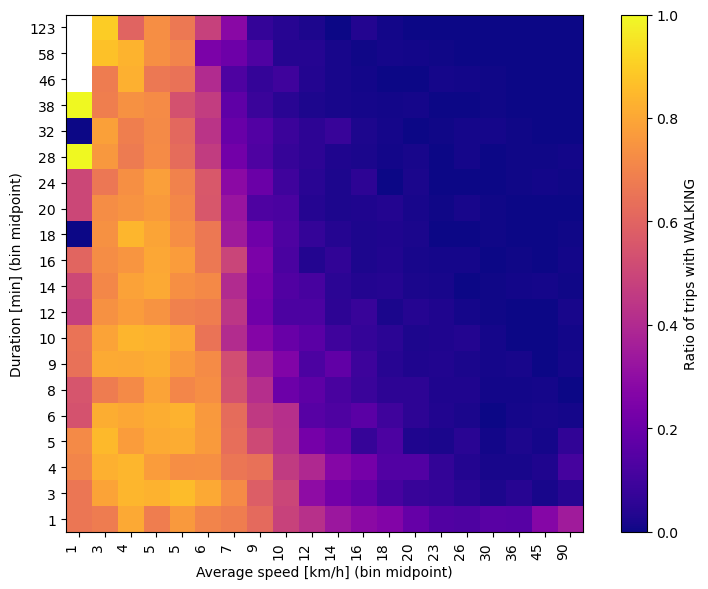}}
\caption{Proportion of trip segments whose recorded mode is walking, by duration and average speed.}
\label{fig:modes-walk}
\end{figure}

Besides the variance of modes, we also examine the choices made by the measured users. 
Figure \ref{fig:modes-walk} shows the frequency of "WALKING" as a trip mode. The data aligns well with the 15-minute city concept regarding average speed, as 5~km/h appears to be a good threshold, except for very short durations. However, the duration behaves differently, with no clear threshold visible. Therefore, it is a conceptual choice whether X-minute cities should aim for a 15- or 20-minute radius.

\subsection{What Makes a Trip ``15 Minutes''? }
Training and evaluating models to predict whether a trip segment falls within the defined 15-minute walking or cycling envelope has revealed several patterns. The models achieve AUC values between 0.75 and 0.85, depending on the target and feature configuration. The same broad groups of features appear among the highest-ranked inputs across the tested model parameters and feature configurations.

\begin{figure}[ht]
\centering
\fbox{\includegraphics[width=0.8\linewidth]{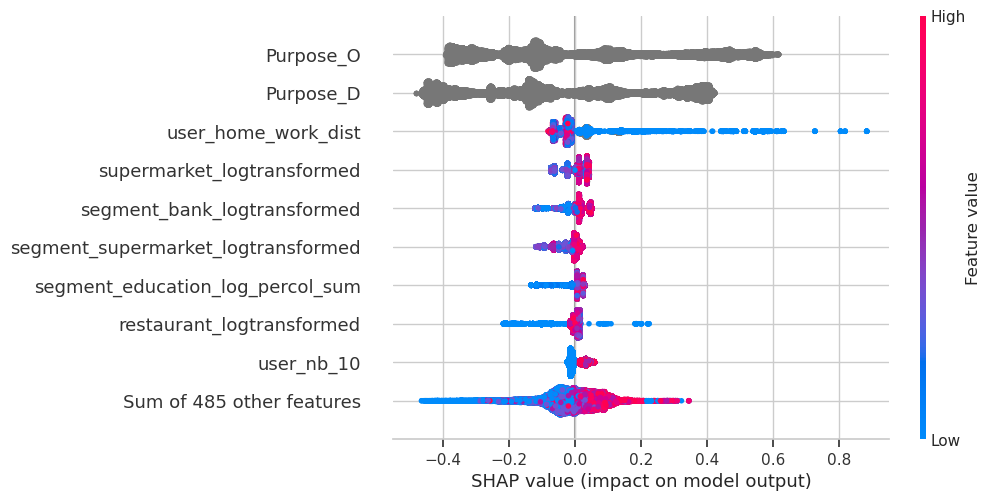}}
\caption{SHAP beeswarm plot for an XGBoost model predicting whether a trip segment falls within the defined 15-minute walking envelope.}
\label{fig:shap-is-segment-15-foot}
\end{figure}

\begin{figure}[ht]
\centering
\fbox{\includegraphics[width=0.8\linewidth]{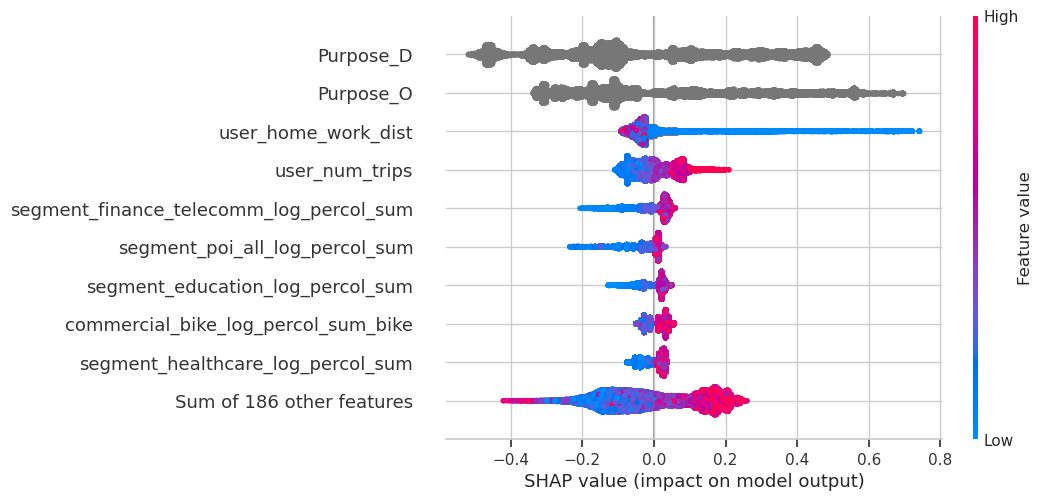}}
\caption{SHAP beeswarm plot for an XGBoost model predicting whether a trip segment falls within the defined 15-minute cycling envelope.}
\label{fig:shap-is-segment-15-bicycle}
\end{figure}

Figure \ref{fig:shap-is-segment-15-foot} illustrates the most important features of the model predicting whether a trip segment falls within the defined 15-minute walking envelope, while Figure \ref{fig:shap-is-segment-15-bicycle} shows the same for the defined 15-minute cycling envelope. 

The purposes of both the trip's origin and destination are among the most influential model inputs for determining whether a trip falls within the 15-minute framework; we add details later with the transport modes. 
Home--work distance is also highly influential, relating to the well-known urban issue of commuting. Additionally, POIs play a significant role: their features are influential when predicting the 15-minuteness of trip segments, particularly when differentiating between cycling-based or walking-based 15-minute trips. For example, the presence of educational POIs within a 15-minute bicycle ride is associated with a higher model prediction for 15-minute cycling trips. Interestingly, the feature $user\_nb\_10$, representing the number of young children in the household, is associated with a higher model prediction for the 15-minute-segment class. For 15-minute cycling segments, a user's total number of trips is an important factor, as these "cycling-sized trips" often imply a higher overall trip frequency. The overall number of POIs relevant to the 15-minute city concept is also a key factor in predicting 15-minuteness.

\subsection{Transport Modes of 15-minute Trips}

Predicting the transport modes of trip segments results in models with AUC values between 0.70 and 0.85. Similar groups of highly ranked features emerge across the tested model configurations. Given this, it is particularly interesting to analyse the key factors associated with transport-mode choices. Among the many noteworthy observations, we highlight two examples: walking and car use.

\begin{figure}[ht]
\centering
\fbox{\includegraphics[width=0.75\linewidth]{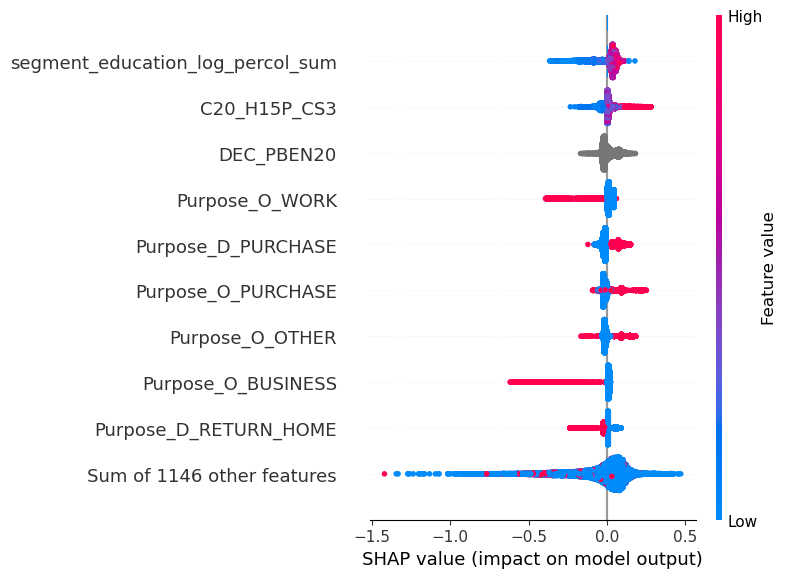}}
\caption{SHAP beeswarm plot for an XGBoost model predicting walking as the main trip mode.}
\label{fig:transport-mode-walking}
\end{figure}

Figure \ref{fig:transport-mode-walking} shows the main factors associated with walking on a trip that is within a 15-minute walking distance. Purposes are one-hot-encoded in this case to better see the details of that category, which is also the most important feature group here. The most important feature is a POI feature: the presence of educational POIs near the segment's starting point. $C20H\_15PC\_S3$ is a sociodemographic feature of the user's home location, indicating higher intellectual professions, which is associated with a higher predicted probability of walking. Similarly, higher income related to $DEC\_PBEN20$ proves to be important. However, walking from work or business origins is not common. Interestingly, going to or coming from a place of purchase increases the chance of walking. One possible explanation is that some lunch-related journeys are labelled as purchases, although this interpretation cannot be verified from the available data.

\subsection{Private Car Use for 15-minute Trips}
Urban car use is a widely debated topic, with ongoing discussions surrounding its environmental, economic, and social impacts. Reducing unnecessary car use is commonly discussed as a component of sustainable urban-mobility policy.

\begin{figure}[ht]
\centering
\fbox{\includegraphics[width=0.6\linewidth]{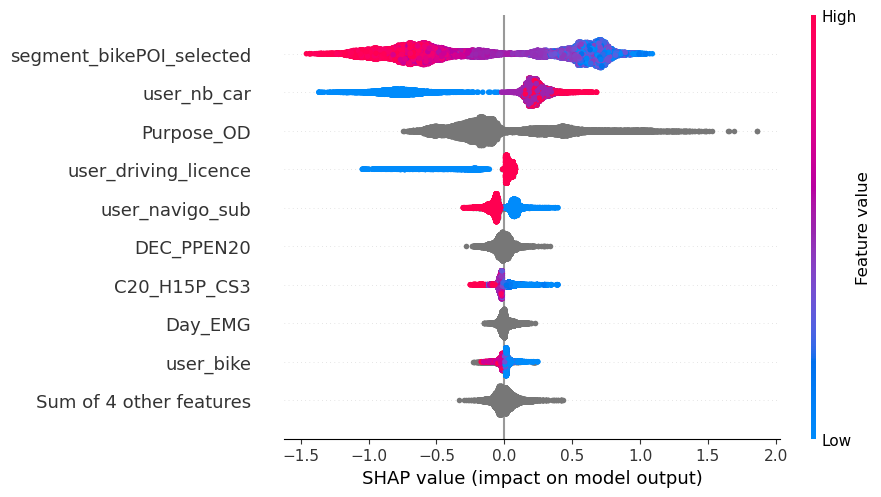}}
\caption{SHAP beeswarm plot for an XGBoost model predicting car use among trips within the defined 15-minute cycling envelope.}
\label{fig:car-use}
\end{figure}

Figure \ref{fig:car-use} provides insight into the factors associated with car use on segments that could be performed on foot or by bicycle too. The XGBoost model evaluated here achieves AUC values between 0.80 and 0.85 across the tested configurations. 

The illustrated model uses the 13 selected features; the most important ones were: 
\begin{itemize}
    \item $segment\_bikePOI\_selected$: weighted number of specific POIs within 15-minute cycling range from the segment start points: education, accessibility, commercial, finance and telecom
    \item $user\_nb\_car$, $user\_bike$, $user\_navigo\_sub$: if the user has a car, bicycle or public-transport subscription
    \item $Purpose\_OD$: the merged origin and destination purposes of the trip
    \item $C20\_H15P\_CS3$: an INSEE statistic for user home location, the number of men aged 15 or older who are classified as "Managers and higher intellectual professions" according to the 2020 census
    \item $DEC\_PPEN20$: an INSEE statistic for user home location,  weighted disposable income of the poorest 20\%
    \item $Day\_EMG$: day of week 
\end{itemize} 

It is important to note that the dataset has limitations: only 4,313 car segments out of a total of 24,743 segments are included, all of which are under 15 minutes and have an average speed of less than 15~km/h. Therefore, the results and patterns extracted from the model should be interpreted with caution, and general conclusions outside the scope of the dataset should be avoided.

The first observation is that POIs near segment start points are important: municipal, educational and commercial POIs are associated with a lower predicted probability of choosing cars as transport mode for these short trip segments. 

Car ownership is associated with a higher predicted probability of car use even for these short, low-speed segments. Driving-licence availability is also associated with a higher predicted probability of car use. Conversely, bicycle ownership is associated with a lower predicted probability of car use.  

\begin{figure}[ht]
\centering
\fbox{\includegraphics[width=0.75\linewidth]{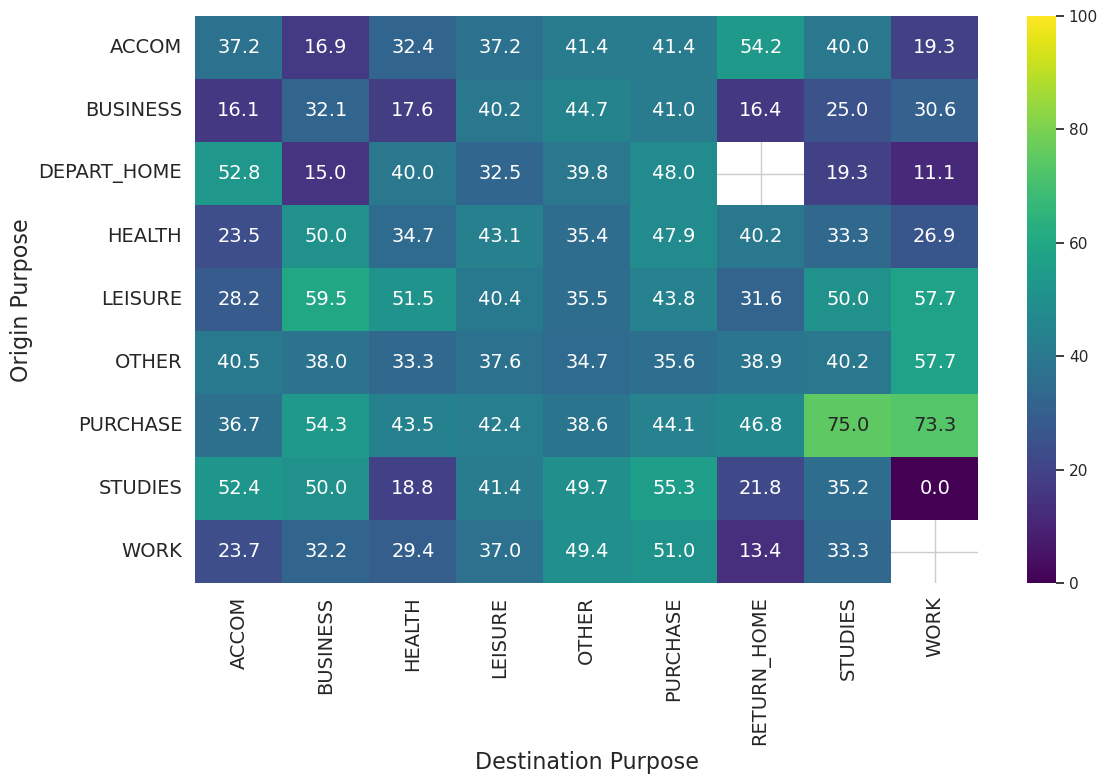}}
\caption{Percentage of maximum 15-minute trips no faster than the average cycling speed, according to the source and destination purposes.}
\label{fig:purpose-15minute}
\end{figure}

\begin{figure}[ht]
\centering
\fbox{\includegraphics[width=0.75\linewidth]{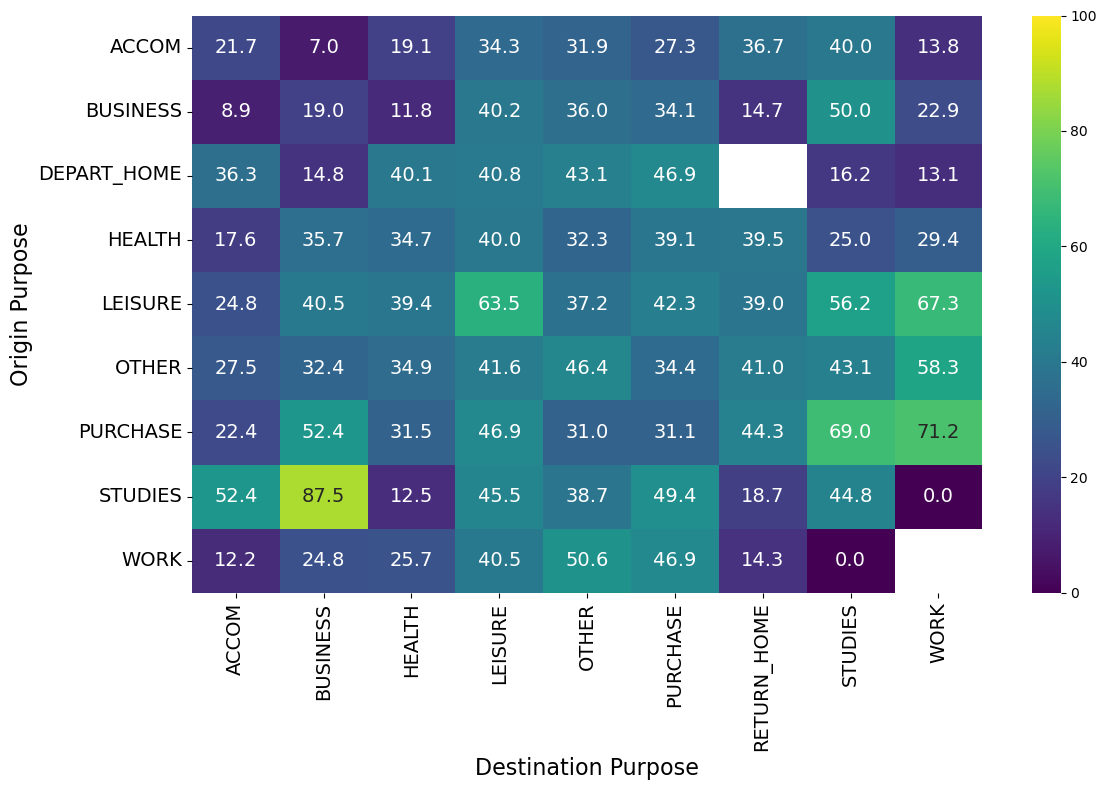}}
\caption{Percentage of trips with walking or bicycle modes, according to the source and destination purposes.}
\label{fig:purpose-walk-bike}
\end{figure}

Purpose of the trip is very important here too, so we examine which purpose pairs imply more walking or cycling segments. Figure \ref{fig:purpose-15minute} provides insight into the relation between 15-minute segments and purposes, while Figure \ref{fig:purpose-walk-bike} shows the actual ratio of trips on foot or by bicycle. Work is notably more related to motorized and longer segments, while purchasing and health-related trips are more walking- or cycling-friendly. 
In our car-use case, the most important cases were the business-business and the purchase-purchase trips; these are also two specific cases in the whole dataset. 

Inspection of selected trees from the XGBoost ensemble illustrates several local feature combinations represented by the model. In gradient boosting, successive trees are fitted to reduce the errors of the current ensemble. Individual trees can therefore illustrate feature combinations used within the fitted model, but they should not be interpreted as independent causal rules. These trees provide interpretable examples that may help formulate policy-relevant hypotheses for further investigation.

\begin{figure}[ht]
\centering
\fbox{\includegraphics[width=0.75\linewidth]{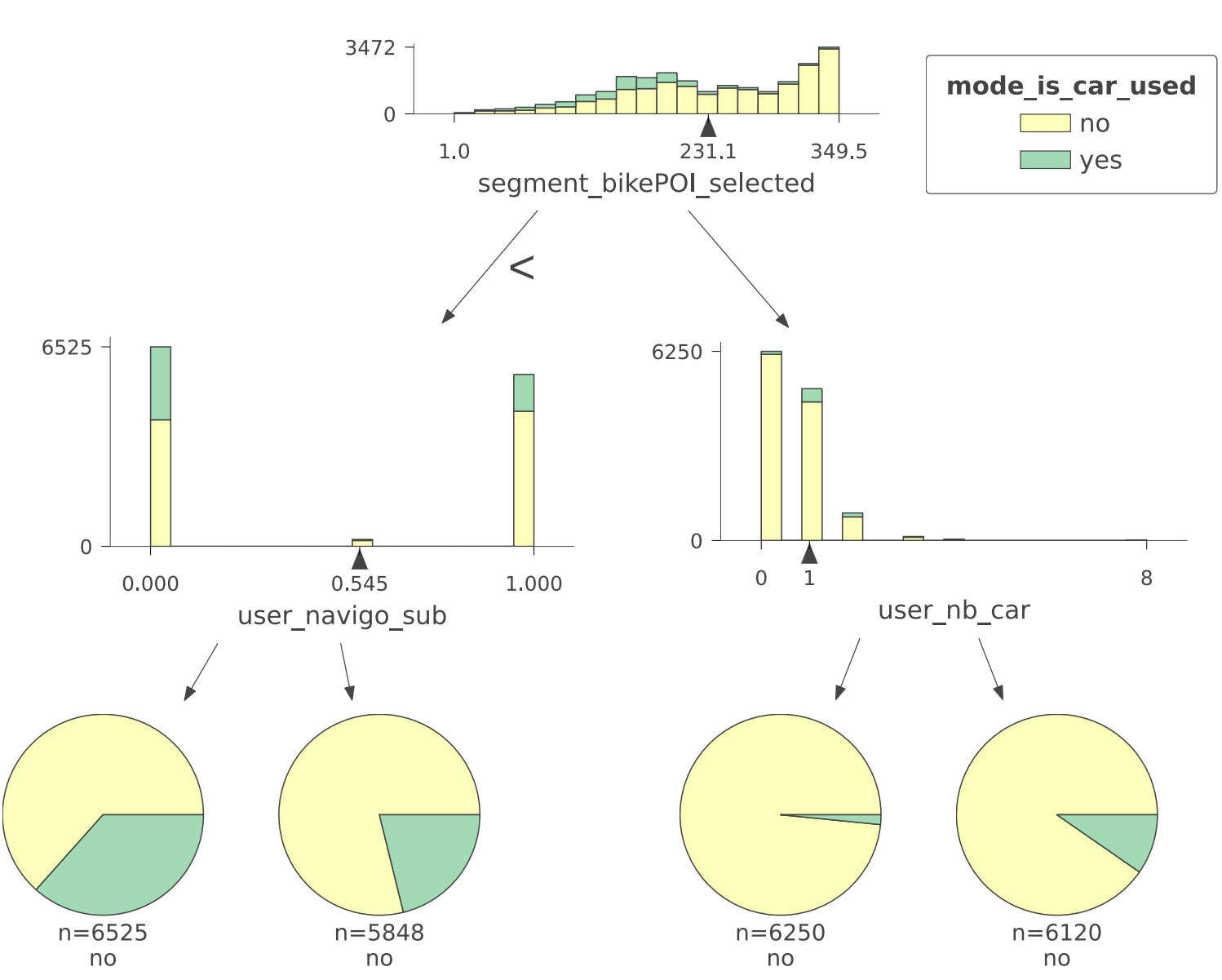}}
\caption{An example decision tree for predicting car use for trip segments no longer than 15 minutes and no faster than the defined cycling-speed threshold.}
\label{fig:tree-car-usage-1}
\end{figure}

\begin{figure}[ht]
\centering
\fbox{\includegraphics[width=0.75\linewidth]{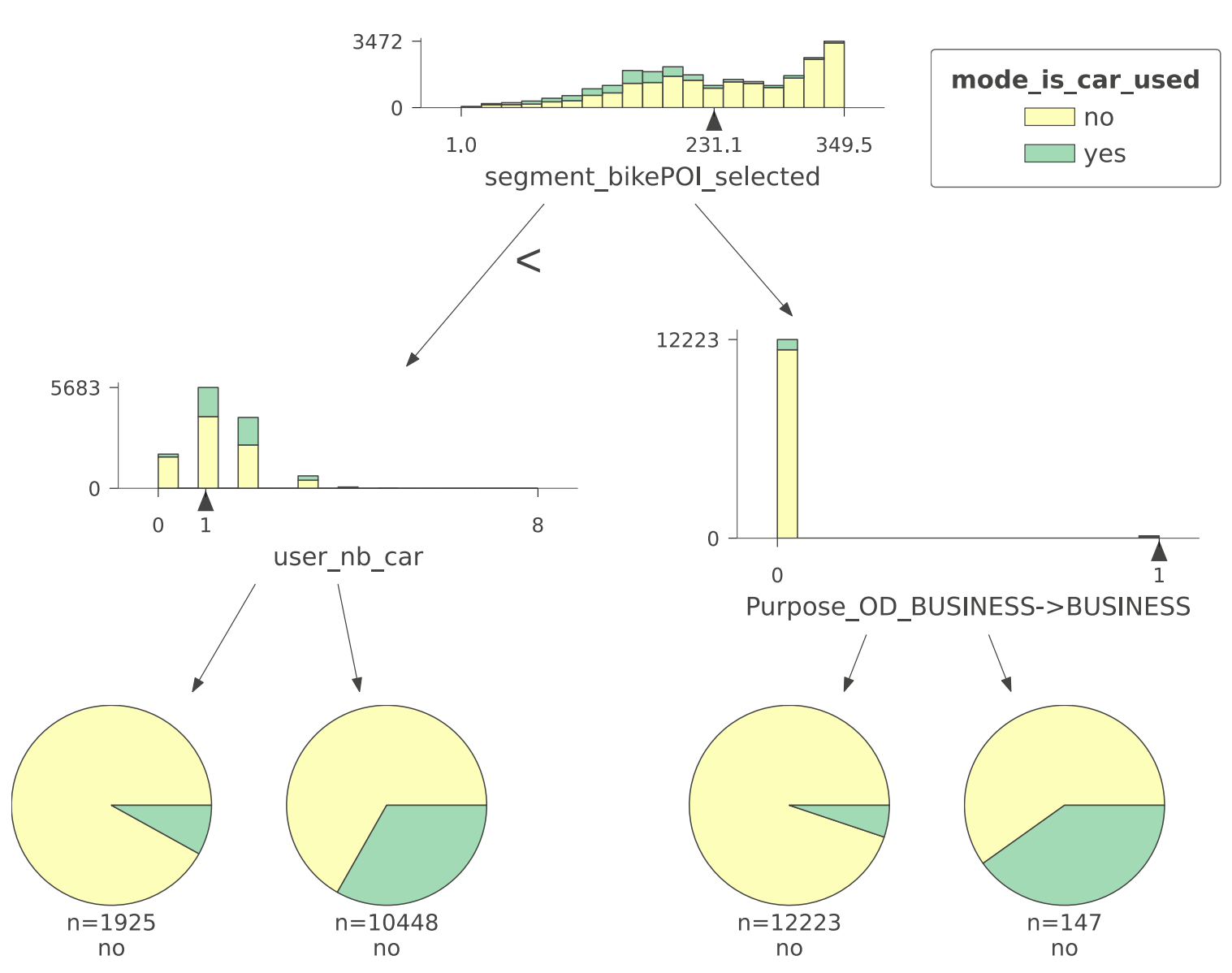}}
\caption{An example decision tree for predicting car use for trip segments no longer than 15 minutes and no faster than the defined cycling-speed threshold.}
\label{fig:tree-car-usage-2}
\end{figure}

Figures \ref{fig:tree-car-usage-1} and \ref{fig:tree-car-usage-2} provide two examples of selected decision trees. The leaves of the trees show the number of occurrences, categorized by car or non-car segments, while the internal nodes separate the groups. The analysis of these trees reveals several key observations:
\begin{itemize}
    \item A high density of POIs is associated with reduced car use.
    \item Conversely, in areas with a low density of POIs, having a public-transport subscription is associated with a lower predicted probability of car use. This suggests that the availability and ease of access to such subscriptions may be linked to transport-mode choice.
    \item Business-to-business trips should be analysed and handled separately, as their transport-mode choice does not appear to be strongly associated with the number of available POIs.
\end{itemize}

Applying asymmetric SHAP can provide deeper insight into model behaviour by moving beyond simple feature correlation. Instead of analysing a flat list of input features and their individual impact on a target variable, this method allows us to examine how feature attributions change under user-specified partial orderings of the input variables. This can be used to explore whether a model attribution is sensitive to hypotheses in which one feature is treated as upstream of another. It does not by itself establish mediation or causal effects. For example, car ownership and driving-licence availability may both be associated with car use. An asymmetric ordering can show how their attributed contributions change when one is treated as upstream of the other, but the ordering must be justified by domain knowledge and is not learned from the observational data.

\begin{figure}[ht]
\centering
\fbox{\includegraphics[width=0.75\linewidth]{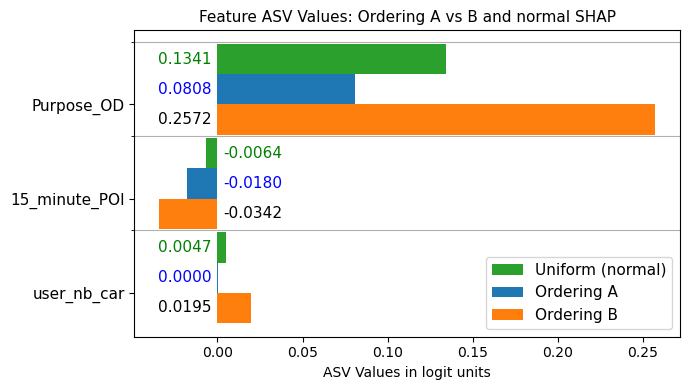}}
\caption{Comparison of conventional SHAP and two asymmetric SHAP orderings for the model predicting private car use on short, low-speed trip segments.}
\label{fig:asymetric-1}
\end{figure}

Figure \ref{fig:asymetric-1} compares the SHAP values of the features of the private car-use model for short, low-speed trips. 
We test whether the sociodemographic features of users precede some other features under the assumed ordering:
\[
\begin{aligned}
&(DEC\_PPEN20, C20\_H15P\_CS3, user\_sex, user\_age) \\
&\quad \longrightarrow (user\_nb\_car, user\_bike, Purpose\_OD, user\_navigo\_sub).
\end{aligned}
\]
"Ordering B" is the opposite direction, while the other features are not included in the ordering. 

$user\_nb\_car$ produces the largest difference with "Ordering A." The increased value, when this feature is treated as an upstream variable under Ordering~A, suggests that its contribution was previously masked by its correlation with other features in the uniform SHAP calculation. It receives a larger attribution when treated as an upstream variable under Ordering~A. A related change is observed for \texttt{user\_navigo\_sub}, although its association with the target is in the opposite direction.

Bicycle ownership, $user\_bike$, however, stays relatively the same. This stability indicates that the attribution assigned to bicycle ownership is relatively insensitive to the tested ordering.

The POI feature is an interesting case. Its Asymmetric Shapley Value (ASV) decreases when applying the orderings, which indicates that its direct attribution decreases when the sociodemographic variables are placed upstream in the assumed ordering. This result is not straightforward to interpret and may reflect correlations among sociodemographic characteristics, POI exposure, and trip origins. Further analysis would be required to determine whether the pattern reflects a substantively meaningful relationship or sensitivity of the attribution method to correlated inputs. These results should be interpreted as sensitivity of model explanations to assumed variable orderings, rather than as identification of causal effects.










\section{Conclusion and Future Work}

The NetMob dataset, enriched with sociodemographic and geographic data, provides a useful basis for studying the 15-minute city concept in Paris using explainable machine-learning methods.

The analysis provides empirical evidence consistent with several central assumptions of the 15-minute city concept. We have identified associations between POI availability and several trip characteristics, including transport-mode choice and short-trip car use. Our focus on "15-minuteness" provides interpretable, policy-relevant hypotheses, with examples of transport-mode selection and private car use demonstrating the applicability of our methodology, which can be used to extract further patterns and insights.

By combining gradient-boosted trees with interpretability frameworks like SHAP and asymmetric SHAP, we were able to identify the most important features associated with trip duration and transport-mode choices. This methodology provides interpretable model explanations that may help identify questions and locations for further urban-policy investigation. For example, areas combining frequent short trips with limited bicycle infrastructure may warrant further investigation when considering measures to promote active mobility. 

As future work, we plan to extend this analysis to examine more complex feature interactions and explanation structures by refining and expanding the analytical framework. Our current findings already highlight significant demographic disparities, such as households with children showing weaker associations with POI proximity and active mobility than some other demographic groups in the analysed data. Interdisciplinary collaboration with urban-planning and sociology experts would be important for evaluating these hypotheses and for ensuring that data-informed interventions do not reinforce existing inequalities.

\clearpage
\bibliography{references}

\end{document}